# Human Activity Recognition using Smartphones

Mayur Sonawane[1] Sahil Dhayalkar[1], Siddesh Waje[1], Soyal Markhelkar[1], Akshay Wattamwar[1],
Prof. Seema C. Shrawne[1]
[1]Department of Computer Engineering & Information Technology,
Veermata Jijabai Technological Institute, Mumbai, India

**Abstract:**
 Human Activity Recognition is a subject of great research today and has its applications in remote healthcare, activity tracking of the elderly or the disables, calories burnt tracking etc. In our project, we have created an Android application that recognizes the daily human activities and calculate the calories burnt in real time. We first captured labeled triaxial acceleration readings for different daily human activities from the smartphone's embedded accelerometer. These readings were preprocessed using a median filter. 42 features were extracted using various methods. We then tested various machine learning algorithms along with dimensionality reduction. Finally, in our Android application, we used the machine learning algorithm and a subset of features that provided maximum accuracy and minimum model building time. This is used for real-time activity recognition and calculation of calories burnt using a formula based on Metabolic Equivalent.

## I. INTRODUCTION

The invention of mobile phones has been a revolutionary one. Ever since the first commercial hand-held mobile phones appeared in 1979, there has been unprecedented growth in the adoption of mobile phone technology. The mobile phones are now used by 75% of the world population as per the statistics provided by the World Bank [1]. People nowadays use their mobile phone for a majority of their time, be it for reading, working out, listening to the music, even as an alarm when we sleep.

With advances in the technology, smartphones are now been equipped with powerful features such as the latest processors that enable multitasking and a variety of sensors [2], in addition to the basic telephony. Therefore, smartphones now make it possible to seamlessly monitor and keep track of our daily activities, learn from them and assist us in making different daily decisions. Such assistive technologies can be of immense use for remote health care, for the elderly, the disabled such as the blind and those with special needs and even a normal person. On the other hand, due to technology making human life easier, lifestyle has turned sedentary and there is no one to look after one's self, which in turn has resulted in health problems, especially obesity, blood pressure, heart attacks, diabetes etc. Having known of this fact, people have now become self-conscious and started doing daily physical activities such as running, fast walking, working out in the gym etc. to burn calories and stay fit.

Human activity recognition is an important yet challenging research area [3], with many applications in healthcare, smart environments, homeland security and aid towards a better technology-driven lifestyle. Computer vision-based techniques and wearable sensors can be used for recognizing daily activities performed by a person, but each has its own disadvantages. For example, we often find people complaining about how uneasy they feel to wear a smartwatch. Cameras are immobile and have storage problems, unless a secondary storage device is available at its disposal, because the video generated consumes a lot of space in the memory, whereas wearable sensors are obtrusive and uncomfortable to a considerable number of people. Smartphones have many hardware sensors embedded in them, such as an accelerometer, gyroscope, orientation sensor etc. [2]. The role of smartphones in recognizing the activities can have several advantages due to easy device portability, and no other requirements for any additional equipment that could be obtrusive and uncomfortable to the user. If automatic activity recognition systems can be built based on an intelligent processing of sensor features on smartphones, it will be a great contribution towards an easier lifestyle, health area, particularly for tracking the number of calories burnt, remote activity monitoring and recognition in elderly care and disability care sector.

Our aim is to create an Android application that recognizes the activities performed by the user (idle, slow walking, normal walking, fast walking, jogging, running, jumping) accurately by sensing the readings provided by the inbuilt sensors in the smartphone (such as an accelerometer or a gyroscope) and then calculate the calories burnt in real time. This can help a user to keep track of the calories he burns on a daily basis and monitor his further activities and food intake. We first gathered our own labeled data generated from the different sensors available on our smartphone. We then performed noise reduction using median filter followed by the feature extraction. Appropriate machine learning and data mining methods need to be developed for processing these sensor signals from smartphones for automatic and intelligent activity recognition. Though there have been several machine learning techniques available, it is not clear, which algorithm can perform better for activity recognition with smartphones. To find the optimum results, in terms of both accuracy and performance on our smartphones, we tested various machine learning algorithms for their accuracy and chose the one which provides maximum accuracy along with less use of resources such as CPU processing.

## II. LITERATURE SURVEY

### A Public domain dataset for human activity recognition using smartphones. [4]

Today's generation of smartphones can be used as an instrument for Human Activity Recognition because they contain inertial sensors such as accelerometers and gyroscopes. Therefore, they can provide an affordable, subtle and unobtrusive solution for monitoring human's daily activity. Using inertial data from smartphone's accelerometer and gyroscope, they have created a dataset which records six human daily activities, namely, sitting, walking, walking upstairs, walking downstairs, standing, laying down.

A set of experiments along with a group of 30 volunteers within age range 19 to 48 were selected to create this dataset. Every person was supposed to wear a waist-mounted smartphone (Samsung Galaxy S II). They performed the six daily activities while placing the smartphone at different locations, namely left side of the belt and according to user's preference. A 5-second gap was maintained between each task. Although the tasks were performed in laboratory conditions the volunteers were instructed to behave freely in order to incorporate some degree of natural behavior in the dataset They collected triaxial (along x, y, and z-axes) linear acceleration and angular velocity signals using the smartphone's accelerometer and gyroscope at a frequency of 50Hz. These signals were then preprocessed for noise reduction using median filter and 3rd order low-pass Butterworth filter with a 20 Hz cutoff frequency.

For computing feature vectors, they used mean, standard deviation, median absolute value, largest value and smallest value in array, signal magnitude area, energy (average sum of squares), interquartile range, correlation coefficients, largest frequency component, autoregression coefficients, frequency signal weighted average, skewness, signal entropy, kurtosis, energy of a frequency interval and angle between two vectors. Finally, they extracted 561 features which correspond to an activity window.

In order to performance assessment, the dataset was randomly partitioned into two sets (70% for training and 30% for testing). Multiclass Support vector machine (SVM) was used as the machine learning algorithm, with 561 features and 6 activities as labels. Training and testing using multiclass SVM gave 90-96% test accuracy. Also, related work on human activity recognition that used special-purpose wearable sensors has shown similar performance (90%-96%). such as a waist-mounted triaxial accelerometer provided an accuracy of 90.8% and a chest-mounted accelerometer provided an accuracy of 93.9%. This concludes that using a smartphone and its sensors for human activity recognition can provide similar (or better) results as compared to special purpose wearable sensors. Moreover, smartphones have the ability to perform human activity recognition in an unobtrusive and less invasive as compared to special purpose sensors.

They have hosted their dataset publicly in UCI Machine Learning Repository by naming it 'Human Activity Recognition Using Smartphones Data Set'.

### SmartPhone Based Data Mining For Human Activity Recognition. [5]

This survey presents novel data analytic scheme for intelligent Human Activity Recognition (AR) using smartphone inertial sensors, with potential applications in automatic assisted living technologies. The researchers mention that although there are enough data and means to collect it using smartphones, there is not enough capacity for automatic decision support. Recent human activity recognition systems aim to identify human actions from the data obtained from sensors and the surrounding environment. This is possible because current smartphones have motion, acceleration or inertial sensors inbuilt in them, and therefore by using the data and information obtained from these sensors, along with appropriate machine learning or deep learning algorithms, human activities can be detected. For experimental validation of their approach, they used the previously mentioned 'Human Activity Recognition Using Smartphones Data Set' from UCI Machine Learning Repository [4].

They propose various data mining approaches for classifying different activities in this work. As the dimensionality of features is very high (561 features), which can severely affect the implementation in real time on smartphone devices, it includes information theory based ranking of features based on its importance or contribution to the machine learning algorithm as the pre-processing step for this purpose. In this approach, the features or attributes are ranked using information gain, and other insignificant features are removed. They carried out extensive experiments with different features ranked by the information gain, and different supervised classifiers such as Naïve Bayes, Decision trees, random forests, random committee (ensemble), and lazy learning (IBk).

Traditional Naïve Bayes classifier performed reasonably well for the large dataset, providing accuracy of 79% accuracy and model building time of 5.76 seconds. However, random forests (similar to an ensemble of decision trees) was better in terms of both accuracy and model building time, with 96.3% accuracy and 14.65 seconds model building time. The other ensemble learning classifiers (such as random committee and random subspace) also gave similar classification accuracy (~96%). On the other hand, unsupervised machine learning algorithms such as k-Means clustering gave poor classification accuracy of with 60% and model build time of 582 seconds. This is due to the fact that sensor readings are oscillating across axes and therefore do not form separated clusters. Also, IBk classifier (based on lazy learning) provided accuracy above 90% for 128 and 256 features.

Therefore, the researchers have examined several learning approaches and found lazy learning, random forests and ensemble learning-based approaches to be promising in terms of activity classification accuracy, model building time.

### Activity Classification with Smartphone Data. [6]

In this paper, the researchers have presented their efforts to develop an effective activity classifier. They drew their data from a prepared dataset. One portion of the dataset contained the raw gyroscope and accelerometer readings, which they called the "raw data." The other portion consisted of vectors that each contain 561 features and represent 2.56 seconds of time. Each vector encodes characteristics such as the triaxial average, minimum and maximum acceleration as well as angular velocity over the given interval, as well as more complex properties such as the Fourier transform and autoregressive coefficients. They called this portion of the data set the "processed data" or "preprocessed data." They used these datasets to train and test the classifiers.

Two approaches were used, first, the testing of pre-processed data and second-time raw data was used. They trained and tested Naive Bayes and Gaussian Discriminant Analysis classifiers on the pre-processed data. They then applied a Hidden Markov Model classifier to GDA classifier, using the activities as the states and the outputs of their GDA as the emissions. The goal in adding an HMM to their GDA classifier was to capture time-based relationships in the data which the static GDA classifier failed to use.

For training and testing the raw data, they first applied the same GDA model that they used previously. They then decided to apply a Hidden Markov Model approach where their states were the activities at a given time step, and using the Baum-Welch algorithm to estimate the probabilities of transitioning between states and the probabilities of state emissions, Further, they used the Viterbi algorithm to predict the series of underlying states for a given test emissions sequence.

### III. SENSORS IN ANDROID

Almost all Android smartphone devices contain various sensors integrated into them. The Android platform supports three broad categories of sensors, namely, motion sensors, environmental sensors and position sensors [7]. Motion sensors measure acceleration forces and rotational forces along three axes and include accelerometers, gravity sensors, gyroscopes, and rotational vector sensors [7]. Environmental sensors measure various environmental parameters, such as ambient air temperature and pressure, illumination, and humidity and include barometers, photometers, and thermometers [7]. Position sensors measure the physical position of a device and include orientation sensors and magnetometers. We are particularly interested in accelerometer [7] sensor, which measures the acceleration force in m/s2 that is applied to a device on all three physical axes (x, y, and z), including the force of gravity.

### IV. DATASET

To gather our own data, we first built a temporary Android application that will record the readings of a smartphone's accelerometer sensor which will be related to the user's movements. The Android application contained options to select the activity and a Start and Stop button. Each group member was instructed to first select the activity option, hold the smartphone in the right hand, press the Start button and do the activity for a minimum of 2 minutes and then press the Stop button. We captured triaxial acceleration from the accelerometer (that is, along with all 3 axes X, Y, and Z) at a frequency of 250Hz. Every group member performed the following activities and the activities were labeled explicitly. The activities we recorded are as follows:
- Idle (Standing)
- Slow walking
- Normal walking
- Fast walking
- Jogging
- Running
- Jumping

Hence, by performing each activity for 3 minutes by all group members, we have a rich amount of labeled dataset. This labeled dataset will serve for training and testing of our model. A snapshot of a part of the dataset is shown in Table 1.

TABLE I
INITIAL DATASET

| accx | accy | accz | activity |
| --- | --- | --- | --- |
| 13.9151 | -5.58328 | -3.60088 | 2 |
| 10.1993 | -2.27928 | -0.61292 | 2 |
| 14.05875 | -7.40287 | 4.884171 | 2 |
| 10.4866 | -10.2951 | 4.405331 | 2 |
| 12.58392 | -7.89129 | -0.67995 | 2 |
| 12.13381 | -4.48195 | -2.94008 | 2 |
| 12.00932 | -8.74362 | -0.96726 | 2 |
| 12.08593 | -4.11803 | -2.15478 | 2 |
| 12.92869 | -4.34787 | 1.704672 | 2 |
| 12.49773 | -9.94073 | 4.309563 | 2 |
| 8.973468 | -3.20823 | 3.65834 | 0 |
| 8.657434 | -3.35188 | 3.476381 | 0 |
| 9.155427 | -3.53384 | 3.409343 | 0 |
| 8.188169 | -3.8403 | 2.920926 | 0 |
| 8.552089 | -3.58173 | 3.284845 | 0 |
| 8.753201 | -3.31358 | 3.093309 | 0 |
| 8.66701 | -3.68707 | 3.275268 | 0 |
| 8.657434 | -3.62961 | 3.102885 | 0 |
| 8.724471 | -3.24654 | 3.303998 | 0 |
| 8.465898 | -3.1795 | 4.118027 | 0 |
| 8.331821 | -3.09331 | 3.878607 | 0 |

accx: accelerometer reading across x-axis.
accy: accelerometer reading across y-axis.
accz: accelerometer reading across z-axis.
activity: The activity performed (0 for idle,
2 for normal walking).

The following graphs (Figure 1, Figure 2, Figure 3) depicts reflect the nature of readings of the accelerometer sensor across all 3 axes:

*X-axis: Activity*
*(0-100     : Idle*
*101-200 : Slow Walking*
*201-300 : Normal Walking*
*301-400 : Fast Walking*
*401-500 : Jogging*
*501-600 : Running*
*601-700 : Jumping)*
*Y-axis: Acceleration in m/s$^2$*

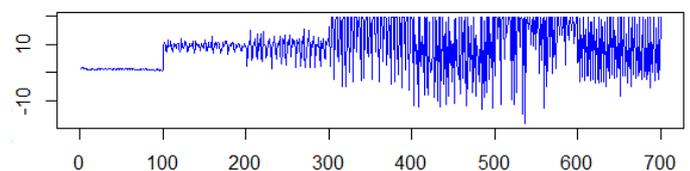

FIGURE 1
ACCELEROMETER READINGS ACROSS X-AXIS

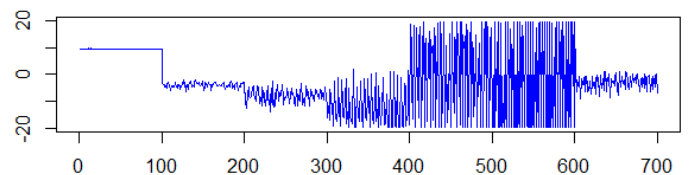

FIGURE 2
ACCELEROMETER READINGS ACROSS Y-AXIS

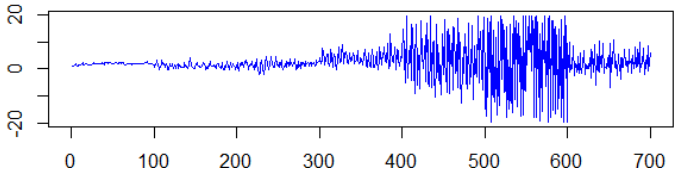

FIGURE 3
ACCELEROMETER READINGS ACROSS Z-AXIS

From the graphs, it can be inferred that activities in which mobile movement is low such as slow walking or idle results are the least amount of fluctuation in the readings while activities in which mobile movement is high such as jogging, running or jumping shows a high amount of fluctuation in the readings. Such fluctuations will be helpful for the machine learning algorithms to classify the readings.

### Noise Reduction

For noise reduction, the readings were preprocessed with a median filter. A graph of some raw readings versus corresponding pre-processed readings using a median filter is depicted in Figure 4.

*X-axis: Activity*
*Y-axis: Acceleration in m/s²*
- Raw readings.
- Pre-processed readings with median filter.

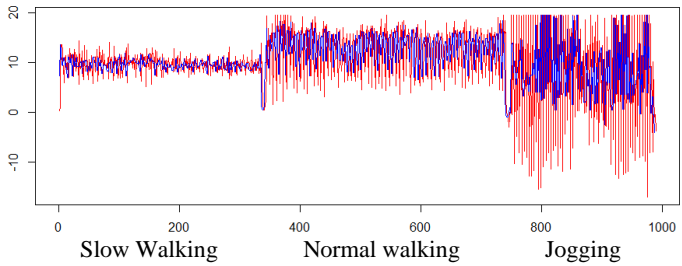

Slow Walking    Normal walking    Jogging

FIGURE 4
ACCELEROMETER READINGS ACROSS Y-AXIS AFTER APPLYING MEDIAN FILTER

It is also important to know whether the median filter increases or decreases the accuracy. Hence we compared both raw and preprocessed (median filter) dataset for their accuracy using various machine learning algorithms. Both the datasets were first randomized and then divided into two datasets, training (70%) and testing (30%). The following table summarizes the testing.

## V. FEATURE EXTRACTION

Feature extraction starts from an initial set of measured data and builds derived values (features) intended to be informative and non-redundant, facilitating the subsequent learning and generalization steps, and in some cases leading to better interpretations. Accelerometer signals are highly fluctuating and oscillatory, which makes it difficult to recognize the underlying patterns using their raw values. Existing HAR systems based on accelerometer data employ statistical feature extraction and, in most of the cases, either time or frequency domain features. Hence we use the following methods to extract features:

- Mean $\bar{x} = \frac{1}{n}\sum_{i=1}^{n} x_i$
- Variance $\sigma^2 = \frac{1}{n}\sum_{i=1}^{n}(x_i - \bar{x})^2$
- Standard deviation $\sigma = \sqrt{\sigma^2} = \sqrt{\frac{1}{n}\sum_{i=1}^{n}(x_i - \bar{x})^2}$
- Interquartile range $IQR = Q_3 - Q_1$
- Energy $e = \frac{1}{n}\sum_{i=1}^{n} x_i^2$
- Kurtosis $\beta_2 = n \frac{\sum_{i=1}^{n}(x_i-\bar{x})^4}{\left(\sum_{i=1}^{n}(x_i-\bar{x})^2\right)^2}$
- Skewness $\gamma_1 = \sqrt{n} \frac{\sum_{i=1}^{n}(x_i-\bar{x})^3}{\left(\sum_{i=1}^{n}(x_i-\bar{x})^2\right)^{3/2}}$

We used a window for every axis that consists of 8 consecutive pre-processed (using a median filter) readings from the accelerometer for that corresponding axis. Hence, for the 3 axes, we have 3 windows. For each window, we calculated the features using the above-mentioned methods to get a single value that against those 8 consecutive readings. From the same window, we derive new 8 values of those 8 consecutive readings using a Fast Fourier Transform. We then calculate the features using the above-mentioned methods to get a single value that against those 8 Fast Fourier transform readings. Hence, we have 42 features (21 features of normal readings of acceleration across x, y z-axis; 21 features of fast Fourier transform of the readings of acceleration across x, y z-axis). They are as follows:

1. *meanaccx*: Mean of 8 preprocessed acceleration values across x-axis.
2. *meanaccy*: Mean of 8 preprocessed acceleration values across y-axis.
3. *meanaccz*: Mean of 8 preprocessed acceleration values across z-axis.
4. *varianceaccx*: Variance of 8 preprocessed acceleration values across x-axis.
5. *varianceaccy*: Variance of 8 preprocessed acceleration values across y-axis.
6. *varianceaccz*: Variance of 8 preprocessed acceleration values across z-axis.
7. *standarddeviationaccx*: Standard deviation of 8 preprocessed acceleration values across x-axis.
8. *standarddeviationaccy*: Standard deviation of 8 preprocessed acceleration values across y-axis.
9. *standarddeviationaccz*: Standard deviation of 8 preprocessed acceleration values across z-axis.
10. *iqraccx*: Inter-quartile range of 8 preprocessed acceleration values across x-axis.
11. *iqraccy*: Inter-quartile range of 8 preprocessed acceleration values across y-axis.
12. *iqraccz*: Inter-quartile range of 8 preprocessed acceleration values across z-axis.
13. *kurtosisaccx*: Kurtosis of 8 preprocessed acceleration values across x-axis.
14. *kurtosisaccy*: Kurtosis of 8 preprocessed acceleration values across y-axis.
15. *kurtosisaccz*: Kurtosis of 8 preprocessed acceleration values across z-axis.
16. *skewnessaccx*: Skewness of 8 preprocessed acceleration values across x-axis.
17. *skewnessaccy*: Skewness of 8 preprocessed acceleration values across y-axis.
18. *skewnessaccz*: Skewness of 8 preprocessed acceleration values across z-axis.
19. *energyaccx*: Energy of 8 preprocessed acceleration values across x-axis.

20. *energyaccy*: Energy of 8 preprocessed acceleration values across y-axis.
21. *energyaccz*: Energy of 8 preprocessed acceleration values across z-axis.
22. *fmeanaccx*: Mean of 8 fast Fourier transform of preprocessed acceleration values across x-axis.
23. *fmeanaccy*: Mean of 8 fast Fourier transform of preprocessed acceleration values across y-axis.
24. *fmeanaccz*: Mean of 8 fast Fourier transform of preprocessed acceleration values across z-axis.
25. *fvarianceaccx*: Variance of 8 fast Fourier transform of preprocessed acceleration values across x-axis.
26. *fvarianceaccy*: Variance of 8 fast Fourier transform of preprocessed acceleration values across y-axis.
27. *fvarianceaccz*: Variance of 8 fast Fourier transform of preprocessed acceleration values across z-axis.
28. *fstandarddeviationaccx*: Standard deviation of 8 fast Fourier transform of preprocessed acceleration values across x-axis.
29. *fstandarddeviationaccy*: Standard deviation of 8 fast Fourier transform of preprocessed acceleration values across y-axis.
30. *fstandarddeviationaccz*: Standard deviation of 8 fast Fourier transform of preprocessed acceleration values across z-axis.
31. *fiqraccx*: Inter-quartile range of 8 fast Fourier transform of preprocessed acceleration values across x-axis.
32. *fiqraccy*: Inter-quartile range of 8 fast Fourier transform of preprocessed acceleration values across y-axis.
33. *fiqraccz*: Inter-quartile range of 8 fast Fourier transform of preprocessed acceleration values across z-axis.
34. *fkurtosisaccx*: Kurtosis of 8 fast Fourier transform of preprocessed acceleration values across x-axis.
35. *fkurtosisaccy*: Kurtosis of 8 fast Fourier transform of preprocessed acceleration values across y-axis.
36. *fkurtosisaccz*: Kurtosis of 8 p fast Fourier transform of reprocessed acceleration values across z-axis.
37. *fskewnessaccx*: Skewness of 8 fast Fourier transform of preprocessed acceleration values across x-axis.
38. *fskewnessaccy*: Skewness of 8 fast Fourier transform of preprocessed acceleration values across y-axis.
39. *fskewnessaccz*: Skewness of 8 fast Fourier transform of preprocessed acceleration values across z-axis.
40. *fenergyaccx*: Energy of 8 fast Fourier transform of preprocessed acceleration values across x-axis.
41. *fenergyaccy*: Energy of 8 fast Fourier transform of preprocessed acceleration values across y-axis.
42. *fenergyaccz*: Energy of 8 fast Fourier transform of preprocessed acceleration values across z-axis.

## VI. FEATURE SELECTION AND TESTING OF MACHINE LEARNING ALGORITHMS

Some features in the dataset might contain redundant or irrelevant information that can negatively affect the recognition accuracy. Selecting a subset of most relevant and useful features helps to reduce computation costs and makes the model simpler. We have generated 42 features. Including all these features will have the following disadvantages:

• Some features will contribute to a negligible extent or may not contribute at all while generating a learning model using a specific machine learning algorithm.

• It will increase the undue processing of the processor and thus use more battery.

• It will increase the response time.

Hence, we will select only a subset of those features that contribute most to the learning. This depends on the type of machine learning algorithm to be used.

The training and testing were carried out using the feature extracted dataset (containing above mentioned 42 features and 1 activity class attribute). This dataset contains 1265 observations/instances. We first randomized the observations. We then divided the dataset into two datasets, namely training set and testing set. The training set constituted 70% and the testing set constituted 30% of the randomized feature extracted dataset. Various machine learning algorithms were tested for accuracy and model building time, using all as well as a set of important features. The selection of machine learning algorithm criteria is higher accuracy, less number of features, and less model building time. Table 3 at the end of the paper summarizes the research carried out.

## VII. HUMAN ACTIVITY CLASSIFICATION (RECOGNITION) IN REAL TIME

Based on the above findings, we have concluded to select Naïve Bayes as the classifier along with the corresponding 10 features for our Android application, since it provides good accuracy with less processing. Also, less processing will be required to compute the 10 features. For real-time activity classification, triaxial accelerometer readings from the smartphone are captured at the same frequency of 250 Hz. The readings are preprocessed using a median filter and the 10 features are generated. Hence, using the trained Naïve Bayes classifier model, we classify (recognize) the current activity in real time. Let the time required for the above processing be x. Similarly, we will classify for 9 more instances. Hence the total time will be 10x. We will now have 10 classified (recognized) real-time activities. After some experiments, it is observed that it requires approximately 2 seconds for the above-mentioned processing. Hence, it is obvious that during that 2 seconds in most cases, the 10 real activities and the 10 classified activities are probably the same. The reason behind selecting 10 activities to be classified at a time is to have a buffer for incorrect classification. However, we will only consider and display that activity which occurs maximum in those 10 classified activities, resulting in the elimination of incorrect classification, if any. Let the time for the whole procedure mentioned above be y. After the activity is recognized, we calculate the calories burnt for that activity.

**Calculation of Calories burnt in real time**

We used the Metabolic Equivalent for Task (MET) for each activity [8]. One MET is defined as 1 kcal/kg/hour and is roughly equivalent to the energy cost of sitting quietly [8]. A MET also is defined as oxygen uptake in ml/kg/min with one MET equal to the oxygen cost of sitting quietly, equivalent to 3.5 ml/kg/min [8]. It is useful in representing the energy cost of activities. Table 2 lists the activities and their MET values [9] [10].

TABLE 2
ACTIVITY AND THEIR CORRESPONDING MET VALUE [9] [10]

| Activity | MET Value |
|---|---|
| Idle | 1.3 |
| Slow Walking | 2.0 |
| Normal Walking | 3.5 |
| Fast Walking | 4.3 |

| Jogging | 7 |
|---|---|
| Running | 14.5 |
| Jumping | 11.8 |

We also require the weight of the person. Let the weight of the person be w. To calculate the calories burnt for the classified value, we use the following formula:

Calories Burnt in seconds = y * w * (MET Value of the activity) / (60 * 60)

## VIII. CONCLUSION

We have successfully carried out the research and implemented our Android application accordingly. We have generated our own labeled raw, preprocessed and feature extracted datasets. We have used Naïve Bayes algorithm and 10 features to classify real-time activities. Also, J48 was a potential candidate for selection, although we did not select it since the tree formed is quite complex. Post implementation testing of real-time activity recognition was a success since the application classified most of the activities correctly. Only when there is a transition between two activities, we have observed incorrect recognition. Since we are using only accelerometer sensor, the battery usage is quite low.

There is still room for improvement. Future work may consider more activities such as squatting, walking upstairs, walking downstairs, etc. We will also address issues such as frequent database updates, battery usage, etc.

TABLE 3

TESTING OF VARIOUS MACHINE LEARNING ALGORITHMS FOR ACCURACY AND MODEL BUILDING TIME

| Sr. No. | Machine Learning algorithm | Number of Features selected | Name of features selected | Accuracy in percentage | Model Building Time required in sec. |
|---|---|---|---|---|---|
| 1.1 | Naïve Bayes | 42 | (all) | 91.3158 | 1.1 |
| 1.2 | Naïve Bayes | 10 | meanaccx meanaccy varianceaccy standarddeviationaccx standarddeviationaccz kurtosisaccy kurtosisaccz skewnessaccx skewnessaccz fkurtosisaccz | 94.2105 | 0* |
| 2.1 | J48 Decision Tree | 42 | (all) | 95.2632 | 0.16 |
| 2.2 | J48 Decision Tree | 11 | meanaccx meanaccy meanaccz varianceaccz skewnessaccx energyaccy energyaccz fmeanaccx fmeanaccz fvarianceaccy fkurtosisaccx | 96.5789 | 0.03 |
| 3.1 | Random Forest | 42 | (all) | 98.4211 | 1.3 |
| 3.2 | Random Forest | 9 | meanaccx meanaccy varianceaccx varianceaccy standarddeviationaccx standarddeviationaccy iqraccx | 97.1053 | 0.78 |
| 4.1 | Bagging | 42 | (all) | 96.8421 | 0.45 |
| 4.2 | Bagging | 8 | *meanaccy*<br>*meanaccz*<br>*varianceaccx*<br>*varianceaccy*<br>*iqraccx*<br>*skewnessaccx*<br>*energyaccz*<br>*fskewnessaccx* | 95.7895 | 0.09 |
| 5.1 | IBk | 42 | *(all)* | 94.4737 | 0* |
| 6.1 | Support Vector Machine | 42 | *(all)* | 95.7895 | 0.25 |
| 6.2 | Support Vector Machine | 14 | *meanaccx*<br>*meanaccy*<br>*meanaccz*<br>*standarddeviationaccx*<br>*standarddeviationaccy*<br>*standarddeviationaccz*<br>*fmeanaccx*<br>*fmeanaccy*<br>*fmeanaccz*<br>*fvarianceaccx*<br>*fiqraccy*<br>*fskewnessaccy*<br>*fskewnessaccz*<br>*fenergyaccy* | 95.5263 | 0.08 |